\pdfoutput=1

\documentclass[11pt]{article}

\usepackage[]{acl}

\usepackage{times}
\usepackage{latexsym}
\usepackage{tabularx}
\usepackage{graphicx}
\usepackage{multirow}
\usepackage{multicol}
\usepackage{booktabs}
\usepackage{xcolor}
\usepackage{tabularx}
\usepackage{enumitem}
\usepackage{soul}
\usepackage{amsmath}

\usepackage[T1]{fontenc}

\usepackage[utf8]{inputenc}

\usepackage{microtype}

\usepackage{xcolor}

\newcommand{\ourData}{\textsc{AllSides}}

\newcommand{\lexrank}{\textsc{Lexrank}}
\newcommand{\bartCNN}{\textsc{BartCNN}}
\newcommand{\pegasusMulti}{\textsc{PegasusMulti}}
\newcommand{\bartMulti}{\textsc{BartMulti}}
\newcommand{\bartNaive}{\textsc{NeuSFT}}
\newcommand{\bartOneStep}{\textsc{NeuS-Title}}

\newcommand{\rougeR}{{\scshape Rouge1-R}}
\newcommand{\bleu}{{\scshape Bleu}}

\newcommand{\pArousal}{$Arousal_{+}$}
\newcommand{\nArousal}{$Arousal_{-}$}
\newcommand{\sumArousal}{$Arousal_{sum}$}
\newcommand{\feqa}{FeQA}
\newcommand{\neuS}{{\scshape NeuS}}

\definecolor{mypink}{RGB}{252, 199, 199}
\definecolor{myorange}{RGB}{255, 231, 173}
\definecolor{mypurple}{RGB}{244, 224, 254}
\definecolor{mygreen}{RGB}{216, 232, 183}
\definecolor{myred}{RGB}{244, 203, 204}
\definecolor{myblue}{RGB}{201, 218, 247}

\title{NeuS: Neutral Multi-News Summarization \\ for Mitigating Framing Bias}

\author{
Nayeon Lee\quad Yejin Bang\quad Tiezheng Yu\quad Andrea Madotto\thanks{$^*$ This work was done when the author was studying at The Hong Kong University of Science and Technology.}\quad Pascale Fung \\
Hong Kong University of Science and Technology \\
\texttt {nyleeaa@connect.ust.hk, pascale@ece.ust.hk}
}

\begin{document}
\maketitle
\begin{abstract}
Media news framing bias can increase political polarization and undermine civil society. The need for automatic mitigation methods is therefore growing. We propose a new task, a \textit{neutral} summary generation from multiple news articles of the varying political leanings
to facilitate balanced and unbiased news reading.
In this paper, we first collect a new dataset, illustrate insights about framing bias through a case study, and propose a new effective metric and model (\bartOneStep) for the task. 
Based on our discovery that title provides a good signal for framing bias, we present \bartOneStep~that learns to neutralize news content in hierarchical order from title to article. Our hierarchical multi-task learning is achieved by formatting our hierarchical data pair (title, article) sequentially with identifier-tokens (``TITLE=>'', ``ARTICLE=>'') and fine-tuning the auto-regressive decoder with the standard negative log-likelihood objective.
We then analyze and point out the remaining challenges and future directions. One of the most interesting observations is that neural NLG models can hallucinate not only factually inaccurate or unverifiable content but also politically biased content.


\end{abstract}

\section{Introduction}
\label{sec:intro}

Media framing bias occurs when journalists make skewed decisions regarding which events or information to cover (information bias) and how to cover them (lexical bias)~\cite{entman2002framing,groeling2013media}. Even if the reporting of the news is based on the same set of underlying issues or facts, the framing of that issue can convey a radically different impression of what actually happened~\cite{gentzkow2006media}.
Since the news media plays a crucial role in shaping public opinion toward various important issues
\cite{de2004effects, mccombs2009news, perse2016media}, bias in media reporting can reinforce the problem of political polarization and undermining civil society rights.  

Allsides.com~\cite{sides2018media} mitigates this problem by displaying articles from various media in a single interface along with an expert-written roundup of news articles. This roundup is a neutral summary for readers to grasp a bias-free understanding of an issue before reading individual articles. Although Allsides fights framing bias, scalability still remains a bottleneck due to the time-consuming human labor needed for composing the roundup.
Multi-document summarization (MDS) models~\cite{lebanoff2018adapting,liu2019hierarchical} could be one possible choice for automating the roundup generation as both multi-document summaries and roundups share a similar nature in extracting salient information out of multiple input articles. Yet the ability of MDS models to provide \textit{neutral} description of a topic issue -- a crucial aspect of the roundup -- remains unexplored.

\begin{figure}
    \centering
    \includegraphics[width=1\linewidth]{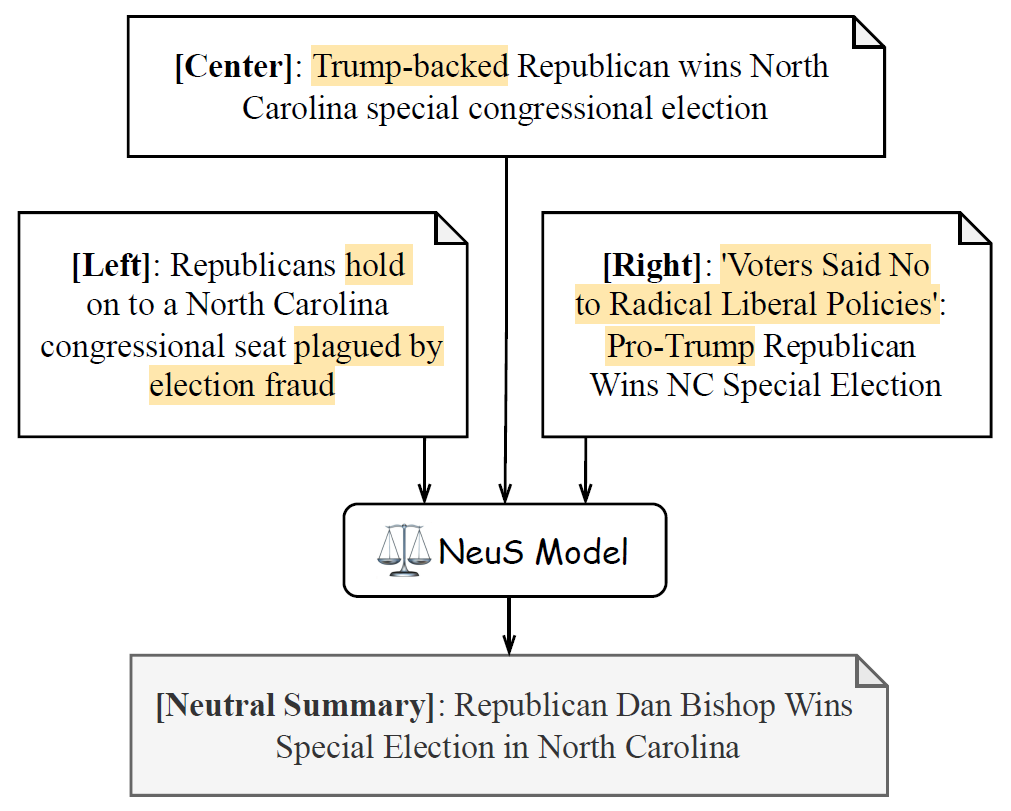}
    \caption{
    Illustration of the proposed task. We want to generate neutral summarization of news articles from varying of political orientations. \colorbox{myorange}{Orange} highlights indicate phrases that can be considered framing bias.}
    \label{fig:intro_example}
\end{figure}


In this work, we fill in this research gap by proposing a task of \underline{Neu}tral multi-news \underline{S}ummarization (\neuS), which aims to generate a framing-bias-free summary from news articles with varying degrees and orientation of political bias (Fig.~\ref{fig:intro_example}).
To begin with, we construct a new dataset by crawling Allsides.com, and investigate how framing bias manifests in the news so as to provide a more profound and more comprehensive analysis of the problem. 
The first important insight from our analysis is a close association between framing bias and the polarity of the text. Grounded on this basis, we propose a polarity-based framing-bias metric that is simple yet effective in terms of alignment with human perceptions.
The second insight is that titles serve as a good indicator of framing bias. Thus, we propose \neuS~models that leverage news titles as an additional signal to increase awareness of framing bias.

Our experimental results provide rich ideas for understanding the problem of mitigating framing bias. Primarily, we explore whether existing summarization models can already solve the problem and empirically demonstrate their shortcomings in addressing the stylistic aspect of framing bias.
After that, we investigate and discover an interesting relationship between framing bias and hallucination, an important safety-related problem in generation tasks. 
We empirically show that the hallucinatory generation has the risk of being not only factually inaccurate and/or unverifiable but also politically biased and controversial. To the best of our knowledge, this aspect of hallucination has not been previously discussed. We thus hope to encourage more attention toward hallucinatory framing bias to prevent automatic generations from fueling political bias and polarization.

We conclude by discussing the remaining challenges to provide insights for future work. We hope our work with the proposed \neuS~task serves is a good starting point to promote the automatic mitigation of media framing bias.

\section{Related Works}
\label{sec:related_works}

\paragraph{Media Bias} 
%
Media bias has been studied extensively in various fields such as social science, economics, and political science. Media bias is known to affect readers' perceptions of news in three main ways: priming, agenda-setting, and framing\footnote{Priming happens when news reporting tells the reader what   context of the event should they evaluate the event in; Agenda-setting is when news reporting tell readers what are the most important problems to think about}\cite{scheufele2000agenda}.
Framing is a broad term that refers to any factor or technique that affect how individuals perceive certain reality or information~\cite{goffman1974frame, entman1993framing, entman2007framing, gentzkow2006media}.
In the context of news reports, framing is about how an issue is characterized by journalists and how readers take the information to form their impression \cite{scheufele2007framing}.
Our work specifically focuses on framing ``bias'' that exists as a form of text in the news. More specifically, we focus on different writing factors such as word choices and the commission of extra information that sway an individual's perception of certain events.

\paragraph{Media Bias Detection}
In natural language processing (NLP), computational approaches for detecting media bias often consider linguistic cues that induce bias in political text~\cite{recasens2013linguistic,yano2010shedding,lee2019team,morstatter2018identifying,lee2019team,hamborg2019illegal,lee2021unifying,bang2021assessing}. For instance, \citeauthor{gentzkow2010drives} count the frequency of slanted words within articles. These methods mainly focus on the stylistic (``how to cover'') aspect of framing bias. However, relatively fewer efforts have been made toward the informational (``what to cover'') aspect of framing bias~\cite{park2011contrasting,fan2019plain}.
Majority of literature doing informational detection are focused on more general factual domain (non-political information) in the name of ``fact-checking''~\cite{thorne2018fever,lee2018improving,lee2021towards,lee2020language}. However, these methods cannot be directly applied to media bias detection because there does not exist reliable source of gold standard truth to fact-check biased text upon.

\paragraph{Media Bias Mitigation}
News aggregation, by displaying articles from different news outlets on a particular topic (e.g., Google News,\footnote{https://news.google.com/} Yahoo News\footnote{https://news.yahoo.com/}), is the most common approach to mitigate media bias~\cite{hamborg2019automated}. 
However, news aggregators require willingness and effort from the readers to be resistant to framing biases and identify the neutral fact from differently framed articles.
Other approaches have been proposed to provide additional information~\cite{laban2017newslens}, such as automatic classification of multiple viewpoints~\cite{park2009newscube}, multinational perspectives~\cite{hamborg2017matrix}, and detailed media profiles~\cite{zhang2019tanbih}. However, these methods focus on providing a broader perspective to readers from an enlarged selection of articles, which still puts the burden of mitigating bias on the readers. Instead, we propose to automatically neutralize and summarize partisan articles to produce a neutral article summary.
\paragraph{Multi-document Summarization}
As a challenging subtask of automatic text summarization, multi-document summarization (MDS) aims to condense a set of documents to a short and informative summary \cite{lebanoff2018adapting}. 
Recently, researchers have applied deep neural models for the MDS task thanks to the introduction of large-scale datasets \cite{liu2018generating,fabbri2019multi}.
With the advent of large pre-trained language models \cite{lewis2019bart,raffel2019exploring}, researchers have also applied them to improve the MDS models, performance~\cite{jin2020multi,pasunuru2021efficiently}. In addition, many works have studied particular subtopics of the MDS task, such as agreement-oriented MDS~\cite{pang2021agreesum}, topic-guided MDS~\cite{cui2021topic} and MDS of medical studies~\cite{deyoung2021ms2}. However, few works have explored generating framing-bias-free summaries from multiple news articles. To add to this direction, we propose the \neuS~task and creates a new benchmark.

\begin{table*}[t]
\centering
\resizebox{0.85\linewidth}{!}{
    \centering
    \begin{tabular}{c|p{15cm}}
        \toprule
         
         \multicolumn{2}{l}{\textbf{Issue A: 
         Trump Put Hold On Military Aid To Ukraine Days Before Call To Ukrainian President}}
        \\ 
        \multicolumn{2}{l}{Left: 
        Trump ordered hold on military aid days before calling Ukrainian president, officials say} \\
        \multicolumn{2}{l}{Right: Trump administration \colorbox{myorange}{\textcolor{black}{claims}} Ukraine aid was \colorbox{myorange}{\textcolor{black}{stalled} over \textcolor{black}{corruption concerns}, \textcolor{black}{decries} media \textcolor{black}{`frenzy'}}}\\
        \multicolumn{2}{l}{Center: Trump Put Hold on Military Aid Ahead of Phone Call With Ukraine’s President}
        \\ \midrule
        
        
        \multicolumn{2}{l}{\textbf{Issue B: Michael Reinoehl appeared to target right-wing demonstrator before fatal shooting in Portland, police say}}
        \\ 
        \multicolumn{2}{l}{Left: \colorbox{myorange}{\textcolor{black}{Suspect} in \textcolor{black}{killing} of right-wing \textcolor{black}{protester} \textcolor{black}{fatally shot} during \textcolor{black}{arrest} }} \\
        \multicolumn{2}{l}{Right: \colorbox{myorange}{Portland's Antifa-supporting \textcolor{black}{gunman} appeared to target \textcolor{black}{victim}, police say}}\\
        \multicolumn{2}{l}{Center: \colorbox{myorange}{Suspect in Patriot Prayer Shooting Killed by Police}}
        \\\midrule 
        
        \multicolumn{2}{l}{\textbf{Issue C: Trump Says the `Fake News Media' Are `the true Enemy of the People'}}
        \\ 
        \multicolumn{2}{l}{Left: President Trump \colorbox{myorange}{renews \textcolor{black}{attacks}} on press as `true \textcolor{black}{enemy} of the people' even as \colorbox{myorange}{CNN receives another \textcolor{black}{suspected bomb}}} \\
        \multicolumn{2}{l}{Right: \colorbox{myorange}{`Great \textcolor{black}{Anger}'} in America caused by `fake news' — \colorbox{myorange}{Trump \textcolor{black}{rips} media for biased reports'}} \\
        \multicolumn{2}{l}{Center: Trump \colorbox{myorange}{blames} 'fake news' \colorbox{myorange}{for country's anger}: 'the true enemy of the people'}
    

        
        \\ \bottomrule
\end{tabular}
}
\caption{Illustration of \colorbox{myorange}{differences in framing} from Left/Right/Center media with examples from \ourData~dataset. We use titles for the analysis of bias, since they are simpler to compare and are representative of the framing bias that exists in the article.}
\label{table:framing_bias_examples}
\end{table*}

\section{Task and Dataset}
\label{sec:task_definition}
\subsection{Task Formulation}
The main objective of \neuS~is to generate a neutral article summary $A_{neu}$ given multiple news articles $A_{0...N}$ with varying degrees and orientations of political bias.
The neutral summary $A_{neu}$ should (i) retain salient information and (ii) minimize as much framing bias as possible from the input articles. 

\subsection{\ourData~Dataset}
\label{section:dataset}
Allsides.com provides access to triplets of news, which comprise reports from left, right, and center American publishers on the same event, with an expert-written neutral summary of the articles and its neutral title. The dataset language is English and mainly focuses on U.S. political topics that often result in media bias. The top-3 most frequent topics\footnote{The full list is provided in the appendix.} are `Elections', `White House', and `Politics'.

We crawl the article triplets\footnote{In some cases, Allsides does not provide all three reportings, and such cases are filtered out.} to serve as the source inputs $\{A_{L},A_{R},A_{C}\}$, and the neutral article summary to be the target output $A_{neu}$ for our task. Note that ``center'' does not necessarily mean completely bias-free~\cite{allsides_2021} as illustrated in Table~\ref{table:framing_bias_examples}. Although ``center'' media outlets are relatively less tied to a particular political ideology, their reports may still contain framing bias because editorial judgement naturally leads to human-induced biases.
In addition, we also crawl the title triplets $\{T_{L},T_{R},T_{C}\}$ and the neutral issue title $T_{neu}$ that are later used in our modeling.

To make the dataset richer, we also crawled other meta-information such as date, topic tags, and media name. In total, we crawled $3,564$ triplets ($10,692$ articles). We use 2/3 of the triplets, which is $2,276$, to be our training and validation set ($80:20$ ratio), and the remaining $1,188$ triples as our test set. We will publicly release this dataset for future research use.


\section{Analysis of Framing Bias}
\label{sec:case_study}
The literature on media framing bias from the NLP community and social science studies provide the definition and types of framing bias \cite{goffman1974frame, entman1993framing, gentzkow2015media, fan2019plain} --- \textit{Informational framing bias} is the biased selection of information (tangential or speculative information) to sway the minds of readers. \textit{Lexical framing bias} is a sensational writing style or linguistic attributes that may mislead readers.
However, the definition is not enough to understand exactly how framing bias manifests in real examples such as, in our case, the \ourData~dataset. We conduct a case-study to obtain concrete insights to guide our design choices for defining the metrics and methodology. 

\subsection{Case-Study Observations}
First, we identify and share the examples of framing bias
in accordance with the literature (Table~\ref{table:framing_bias_examples}).

\paragraph{Informational Bias} 
This bias exists dominantly in the form of ``extra information'' on top of the salient key information about an issue that changes the overall impression of it.
For example, in Table~\ref{table:framing_bias_examples}, when reporting about the hold put on military aid to Ukraine (Issue A), the right-leaning media reports the speculative claim that there were ``corruption concerns'' and tangential information ``decries media `frenzy''' that amplifies the negative impression of the issue.
Sometimes, media with different political leanings report additional information to convey a completely different focus
on the issue. For Issue C, left-leaning media implies that Trump's statement about fake news has led to ``CNN receiving another suspected bomb'', whereas right-leaning media implies that the media is at fault by producing ``biased reports''. 

\paragraph{Lexical Bias} 
This bias exists mainly as biased word choices that change the nuance of the information that is being delivered.
For example, in Issue B, we can clearly observe that two media outlets change the framing of the issue by using different terms ``suspect'' and ``gunman'' to refer to the shooter, and ``protester'' and ``victim'' to refer to the person shot. 
Also, in Issue A, when one media outlet uses ``(ordered) hold'', another media uses ``stalled'', which has a more negative connotation. 

\subsection{Main Insights from Case-Study}
Next, we share important insights from the case study observation that guide our metric and model design.

\paragraph{Relative Polarity}
Polarity is one of the commonly used attributes in identifying and analyzing framing bias~\cite{fan2019plain,recasens2013linguistic}.
Although informational and lexical bias is conceptually different, both are closely associated with polarity changes of concept, i.e., positive or negative, to induce strongly divergent emotional responses from the readers~\cite{hamborg2019illegal}. 
Thus, polarity can serve as a good indicator of framing bias. 

However, we observe that the polarity of text must be utilized with care in the context of framing bias. \textit{It is the relative polarity that is meaningful to indicate the framing bias, not the absolute polarity}. To elaborate, if the news issue itself is about tragic events such as ``Terror Attack in Pakistan'' or ``Drone Strike That Killed 10 people'', then the polarity of neutral reporting will also be negative. 

\paragraph{Indicator of Framing} We discover that the news title is very representative of the framing bias that exist in the associated articles -- this makes sense because the title can be viewed as a succinct overview of the content that follows\footnote{There can be exceptions for the non-main-stream media that use clickbait titles.}. For instance, in Table~\ref{table:generation_examples} the source input example, the right-leaning media's title, and article are mildly mocking of the ``desperate'' democrats' failed attempts to take down President Trump. In contrast, the left-leaning media's title and an article show a completely different frame -- implying that many investigations are happening and there is ``possible obstruction of justice, public corruption, and other abuses of power.'' 

\section{Metric}
\label{sec:metric} 
We use three metrics to evaluate summaries from different dimensions. For framing bias, we propose a polarity-based metric based on the careful design choices detailed in \S\ref{subsec:metric_bias}. For evaluating whether the summaries retain salient information, we adopt commonly used information recall metrics (\S\ref{subsec:metric_recall}). In addition, we use a hallucination metric to evaluate if the generations contain any unfaithful hallucinatory information because the existence of such hallucinatory generations can make the summary fake news (\S\ref{subsec:hallu}).

\subsection{Framing Bias Metric}
\label{subsec:metric_bias}
\subsubsection{Design Consideration}
Our framing bias metric is developed upon the insights we obtained from our case study in \S\ref{sec:case_study}.

First of all, we propose to build our metric based on the fact that framing bias is closely associated with polarity.
Both model-based and lexicon-based polarity detection approaches are options for our work, and we leverage the latter for the following reasons: 1) There is increasing demand for interpretability in the field of NLP~\cite{belinkov2020interpretability, sarker2019interpretable}, and the lexicon-based approach is more interpretable (provides token-level human interpretable annotation) compared to black-box neural models. 
2) In the context of framing bias, distinguishing the subtle nuance of words between synonyms is crucial (e.g., dead vs. murdered). 
The lexicon-resource provides such token-level fine-grained scores and annotations, making it useful for our purpose.

Metric calibration is the second design consideration, and is motivated by our insight into the relativity of framing bias.
The absolute polarity of the token itself does not necessarily indicate framing bias (i.e., the word ``riot'' has negative sentiment but does not always indicate bias), so it is essential to measure the relative degree of polarity.
Therefore, calibration of the metric in reference to the neutral target is important. 
Any tokens existing in the neutral target will be ignored in bias measurement for the generated neutral summary. For instance, if ``riot'' exists in the neutral target, it will not be counted in bias measurement through calibration.

\subsubsection{Framing Bias Metric Details}

For our metric, we leverage Valence-Arousal-Dominance 
(VAD)~\cite{mohammad2018obtaining} dataset which has a large list of lexicons annotated for valence ($v$), arousal ($a$) and dominance ($d$) scores. Valence, arousal, and dominance represent the direction of polarity (positive, negative), the strength of the polarity (active, passive), and the level of control (powerful, weak), respectively. 

Given the neutral summary generated from the model $\hat{A}_{neu}$, our metric is calculated using the VAD lexicons in the following way: 
    

\begin{enumerate}[noitemsep]
    \item Filter out all the tokens that appear in neutral target $A_{neu}$ to obtain set of tokens \textit{unique} to $\hat{A}_{neu}$. This ensures that we are measuring the \textit{relative} polarity of $\hat{A}_{neu}$ in reference to the neutral target $A_{neu}$ -- results in calibration effect. 
    
    \item Select tokens with either positive valence ($v > 0.65$) or negative valence ($v < 0.35$) to eliminate neutral words (i.e., stopwords and non-emotion-provoking words) -- this step excludes tokens that are unlikely to be associated with framing bias from the metric calculation.
    
    \item Sum the arousal scores for the identified positive and negative tokens from Step 2 -- positive arousal score (\pArousal) and negative arousal score (\nArousal). We intentionally separate the positive and negative scores for finer-grained interpretation. We also have the combined arousal score (\sumArousal=\pArousal+\nArousal) for a coarse view. 

    \item Repeat for all \{$A_{neu}$, $\hat{A}_{neu}$\} pairs in the testset, and calculate the average scores to use as the final metric. We report these scores in our experimental results section (\S\ref{sec:results_and_analysis}).
\end{enumerate}

\textit{In essence, our metric approximates the existence of framing bias by quantifying how intensely aroused and sensational the generated summary is in reference to 
the target neutral reference.}
We publicly release our metric code for easy use by other researchers\footnote{https://github.com/HLTCHKUST/framing-bias-metric}.

\subsubsection{Human Evaluation}

To ensure the quality of our metric, we evaluate the correlation between our framing bias metric and human judgement.
We conduct A/B testing\footnote{Please refer the appendix for more detail of the A/B testing} where the annotators are given two generated articles about an issue, one
with a higher \sumArousal~score and the other with a lower score. Then, annotators are asked to select the more biased article summary.
When asking which article is more ``biased'', we adopt the question presented by \citeauthor{spinde2021you}
We also provide examples and the definition of framing bias for a better understanding of the task. We obtain three annotations each for $50$ samples and select those with the majority of votes.

A critical challenge of this evaluation is in controlling the potential involvement of the annotators' personal political bias.
Although it is hard to eliminate such bias completely, we attempt to avoid it by collecting annotations from those indifferent to the issues in the test set. Specifically, given that our test set mainly covers US politics, we restrict the nationality of annotators to non-US nationals who view themselves bias-free towards any US political parties.



After obtaining the human annotations from A/B testing, we also obtain automatic annotation based on the proposed framing bias metric score, where the article with a higher \sumArousal~is chosen to be the more biased generation.
The Spearman correlation coefficient between human-based and metric-based annotations is 0.63615 with a p-value < 0.001, and the agreement percentage $80\%$. These values indicate that the association between the two annotations is statistically significant, suggesting that our metric provides a good approximation of the existence of framing bias.

\subsection{Salient Info}
\label{subsec:metric_recall}
The generation needs to retain essential/important information while reducing the framing bias. Thus, we also report {\scshape Rouge}~\cite{lin2004rouge} and \bleu~\cite{papineni2002bleu} between the generated neutral summary, $\hat{A}_{neu}$, and human-written summary, $A_{neu}$. Note that {\scshape Rouge} measures the recall (i.e., how often the n-grams in the human reference text appear in the machine-generated text) and \bleu~measures the precision (i.e., how often the n-grams in the machine-generated text appear in the human reference text).
The higher the \bleu~and \rougeR~score, the better the essential information converges. In our results, we only report Rouge-1, but Rouge-2 and Rouge-L can be found in the appendix. 


\subsection{Hallucination Metric}
\label{subsec:hallu}
Recent studies have shown that neural sequence models can suffer from the hallucination of additional content not supported by the input~\cite{reiter2018structured,wiseman-etal-2017-challenges,nie2019simple, maynez-etal-2020-faithfulness,pagnoni2021understanding,ji2022survey}, consequently adding factual inaccuracy to the generation of NLG models.
Although not directly related to the goal of \neuS, we evaluate the hallucination level of the generations in our work.
We choose a hallucination metric called FeQA~\cite{durmus-etal-2020-feqa} because it is one of the publicly available metrics known to have a high correlation with human faithfulness scores. This is a question-answering-based metric built on the assumption that the same answers will be derived from hallucination-free generation and the source document when asked the same questions.

\begin{table*}
\centering
\resizebox{0.85\linewidth}{!}{
\begin{tabular}{lccc|cc|c}
\toprule
 & \multicolumn{3}{c|}{\textbf{Avg. Framing Bias Metric}} & \multicolumn{2}{c|}{\textbf{Salient Info}} & \textbf{Hallucination} \\ \cmidrule{2-7} 
\multirow{-2}{*}{\textbf{Models}} & \pArousal $\downarrow$ & \nArousal $\downarrow$ & \sumArousal $\downarrow$ & \bleu $\uparrow$ & \rougeR $\uparrow$ & \feqa $\uparrow$ \\ \midrule
All Source input & 6.76 & 3.64 & 10.40 & 8.27 & 56.57\% & - \\ \midrule
\lexrank & 3.02 & 1.74 & 4.76 & \textbf{12.21} & 39.08\% & 53.44\% \\
\bartCNN & 2.09 & 1.23 & 3.32 & 10.49 & 35.63\% & 58.03\% \\
\pegasusMulti & 5.12 & 2.39 & 7.51 & 6.12 & \textbf{44.42\%} & 22.24\% \\
\bartMulti & 5.94 & 2.66 & 8.61 & 4.24 & 35.76\% & 21.06\% \\
\bartNaive & 1.86 & 1.00 & 2.85 & 11.67 & 35.11\% & \textbf{58.50\%} \\ \midrule
\bartOneStep & \textbf{1.69} & \textbf{0.83} & \textbf{2.53} & 12.05 & 36.07\% & 45.95\%
\\ \bottomrule
\end{tabular}}
\caption{Experimental results for \ourData~test set. We provide the level of framing bias inherent in ``source input'' from the \ourData~test set to serve as a reference point for the framing bias metric.
For framing bias metric, the \textit{lower} number is the better ($\downarrow$). For other scores, the \textit{higher} number is the better ($\uparrow$).}
\label{table:main_result}
\end{table*}

\section{Models and Experiments\footnote{Experimental details are provided in the appendix for reproducibility.}}
\label{sec:experiment}
\subsection{Baseline Models}
Since one common form of framing bias is the reporting of extra information (\S\ref{sec:case_study}), summarization models, which extract commonly shared salient information, may already generate a neutral summaries to some extent.
To test this, we conduct experiments using the following baselines.


\begin{itemize}[noitemsep]
    \item \lexrank~\cite{erkan2004lexrank}: an extractive single-document summarization (SDS) model that extracts representative sentences that hold information common in both left- and right-leaning articles.
     
    \item \bartCNN: an abstractive SDS model that fine-tunes BART-large~\cite{lewis2019bart}, with 406M parameters, using the CNN/DailyMail~\cite{hermann2015teaching} dataset.
    
    \item \bartMulti: a multi-document summarization (MDS) model that fine-tunes BART-large using Multi-News~\citep{fabbri2019multi} dataset.
    
    \item \pegasusMulti: an MDS model that fine-tunes Pegasus-base~\cite{zhang2019pegasus}, with 568M parameters, using the Multi-News dataset.
\end{itemize}

\noindent Since the summarization models are not trained with in-domain data, we provide another baseline model trained with in-domain data for a full picture. 

\begin{itemize}[noitemsep]
    \item \bartNaive: a baseline that fine-tunes the BART-large model using \ourData.
\end{itemize}

\subsection{Our \neuS~Models (\bartOneStep)}
We design our models based on the second insight from the case study (\S\ref{sec:case_study}) - the news title serves as an indicator of the framing bias in the corresponding article. We hypothesize that it would be helpful to divide-and-conquer by neutralizing the the title first, then leveraging the ``neutralized title'' to guide the final neutral summary of the longer articles.

Multi-task learning (MTL) is a natural modeling choice because two sub-tasks are involved -- title-level and article-level neutral summarization. 
Meanwhile, we also have to ensure a hierarchical relationship between the two tasks in our MTL training because article-level neutral summarization leverages the generated neutral title as an additional resource.
We use a simple technique to do hierarchical MTL 
by formatting our hierarchical data pair (title, article) in a single natural language text with identifier-tokens (``Title=>'', ``Article=>''). 
This technique allows us to optimize for both title and article neutral summarization tasks easily
by optimizing for the negative log-likelihood of the single target Y.
The auto-regressive nature of the decoder also ensures the hierarchical relationship between the title and article.


We train BART's autoregressive decoder to generate the target text $Y$ formatted as follows:
\begin{align*}
 \texttt{TITLE}\Rightarrow T_{neu}. \  \texttt{ARTICLE}\Rightarrow A_{neu},
\end{align*}
where $T_{neu}$~and $A_{neu}$ denote the neutral title and neutral article summary.

The input $X$ to our BART encoder is formatted similarly to the target text $Y$:
\begin{align*}
 &\texttt{TITLE}\Rightarrow T_{L}. \  \texttt{ARTICLE}\Rightarrow A_{L}. [SEP] \\
 &\texttt{TITLE}\Rightarrow T_{C}. \  \texttt{ARTICLE}\Rightarrow A_{C}. [SEP] \\
 &\texttt{TITLE}\Rightarrow T_{R}. \  \texttt{ARTICLE}\Rightarrow A_{R}, \
\end{align*}
where $T_{L/C/R}$~and $A_{L/C/R}$ denote the title and article from left-wing, center, and right-wing media, and [SEP] denotes the special token that separates different inputs. Note that the order of left, right, and center are randomly shuffled for each sample to discourage the model from learning spurious patterns from the input. 

\begin{table*}[t]
\centering
\small
\resizebox{\linewidth}{!}{
    \centering
    \begin{tabular}{p{15cm}}
        \toprule
        \textbf{SOURCE:}
        \textbf{<Left>} \textbf{Title:} Here Are The 81 People And Entities Close To Trump Democrats Are Investigating.
        \textbf{Article:} Democrats on the House Judiciary Committee on Monday sent document requests to 81 agencies, entities and individuals close to President Donald Trump as part of a broad investigation into possible obstruction of justice, public corruption and other abuses of power. The list includes Trump’s sons, Eric Trump and Donald Trump Jr., as well as his son-in-law, Jared Kushner. 
        
        \textbf{<Center>} \textbf{Title:} House Panel Requests Documents From Associates of Trump.
        \textbf{Article:} House Democrats intensified their investigations into President Trump and his associates Monday, demanding records from more than 80 people and organizations related to his business dealings, interactions with the Justice Department and communications with Russian President Vladimir Putin.
        
        \textbf{<Right>} \textbf{Title:} Dems Continue Their Assault on The Trump Administration By Launching Another Probe.
        \textbf{Article:} Democrats are desperate to take down President Donald Trump. The Russia probe has proven to be ineffective and, quite frankly, a waste of time and taxpayer money. They didn't find what they wanted so now they're launching another probe.
        \\\midrule
        \textbf{TARGET:} House Democrats launched a broad probe into President Trump on Monday, requesting documents from 81 agencies and individuals as they investigate his business dealings, interactions with Russia, and possible obstruction of justice.
        \\ \midrule \midrule
        \textbf{Lexrank:} Democrats are \textcolor{red}{desperate} to take \textcolor{red}{down} President Donald Trump. The Russia probe has proven to be \textcolor{red}{ineffective} and, quite frankly, a \textcolor{red}{waste} of time and taxpayer money.
        \\ \midrule
        \textbf{\bartNaive:} The Russia probe has proven to be \textcolor{red}{ineffective} and, quite frankly, a \textcolor{red}{waste} of time and taxpayer money. 
        \\ \midrule
        \textbf{\bartOneStep:} TITLE=> House Panel Requests Documents. ARTICLE=> The House Select Committee on Intelligence has requested documents from 81 people and entities \textcolor{red}{close} to President Trump, including his sons Eric and Donald Trump Jr., as well as Jared Kushner. 
\\ \bottomrule
\end{tabular}
}
\caption{Generation examples for analysis purposes. \textcolor{red}{Red} highlights the tokens identified by VAD lexicons. Refer to the appendix for more examples.}
\label{table:generation_examples}
\end{table*}

\begin{table*}[t]
\centering
\small
\resizebox{\linewidth}{!}{
    \centering
    \begin{tabular}{p{15cm}}
        \toprule
        \textbf{SOURCE:} ... President Trump on Saturday blasted what he called the ``phony'' BuzzFeed story and the mainstream media’s coverage of it....
        \\ \midrule
        \textbf{MDS Hallucination:} president trump on sunday slammed what he called called a ``phony'' story by the \colorbox{myorange}{``dishonest'' and ``fake news'' news outlet} in a series of tweets. ... \colorbox{myorange}{``the fake news media is working overtime to make} \colorbox{myorange}{this story look like it is true,'' trump tweeted. ``they are trying to make it look like the president is trying to hide} \colorbox{myorange}{something, but it is not true!''}
\\ \bottomrule
\end{tabular}
}
\caption{Illustration of \colorbox{myorange}{hallucinatory framing bias} from MDS models and the corresponding ``most relevant source snippet'' from the source input. Refer to the appendix for more examples with full context.}
\label{table:hallucination_examples}
\end{table*}

\section{Results and Analysis}
\label{sec:results_and_analysis}
In this section, we point out noteworthy observations from the quantitative results in Table~\ref{table:main_result} along with insights obtained through qualitative analysis. Table~\ref{table:generation_examples} shows generation examples that are most representative of the insights we share.\footnote{More examples are provided in the appendix.}

\subsection{Main Results}
Firstly, summarization models can reduce the framing bias to a certain degree (drop in \sumArousal~score from 10.40 to 4.76 and 3.32 for \lexrank~and \bartCNN). This is because informational framing bias is addressed when summarization models extract the most salient sentences, which contain common information from the inputs. 
However, summarization models, especially \lexrank~cannot handle the lexical framing bias, as shown in Table~\ref{table:generation_examples}. 
Moreover, if we further observe the results of \lexrank, it is one of the best performing models in terms of \rougeR~($39.08\%$), the standard metric for summarization performance, but not in terms of the framing bias metric. This suggests that \textit{having good summarization performance (\rougeR) does not guarantee that the model is also neutral} -- i.e., the requirement for summaries to be neutral adds an extra dimension to the summarization task.

Secondly, one interesting pattern that deserves attention is that only the single-document summarization models (\bartCNN~and \lexrank) reduced framing bias well, not the multi-document summarization models (\pegasusMulti~and \bartMulti). This is rather surprising because our task setup is more similar to MDS than SDS.
One of the major contributors to high bias in the MDS models is probably the hallucination because MDS models portray drastically poor hallucination performance than all the other models (both the MDS models \pegasusMulti~and \bartMulti~achieve 22.24\% and 21.06\%, when most of the other models achieve over 50\%).\footnote{Note that 22.24\% and 21.06\% are already high FeQA scores, however, a comparatively low score in reference.} This suggests that the framing bias of MDS models may be related to the hallucination of politically biased content. We investigate into this in the next subsection (\S\ref{subsection:further_analysis}).

Thirdly, although summarization models help reduce the framing bias scores, we, unsurprisingly, observe a more considerable bias reduction when training with in-domain data. \bartNaive~shows a further drop across all framing bias metrics without sacrificing the ability to keep salient information. However, we observe that \bartNaive~often copies directly without any neutral re-writing -- e.g., the \bartNaive~example shown in Table~\ref{table:generation_examples} is a direct copy of the sentence from the input source. 

Lastly, we can achieve slightly further improvement with \bartOneStep~across all metrics except the \feqa~score.
This model demonstrates a stronger tendency to 
paraphrase rather than directly copy, and has comparatively more neutral framing of the issue. 
As shown in Table~\ref{table:generation_examples}, when \lexrank~and \bartNaive~are focused on the ``ineffectiveness of Russia probe'', the TARGET and \bartOneStep~focus on the start of the investigation with the request for documents. 
\bartOneStep~also generate a title with a similar neutral frame to the TARGET, suggesting this title generation guided the correctly framed generation. 

\subsection{Further Analysis and Discussion}
\label{subsection:further_analysis}
\paragraph{Q: Is hallucination contributing to the high framing bias in MDS models?} Through qualitative analysis, we discovered the MDS generations were hallucinating politically controversial or sensational content that did not exist in the input sources. This is probably originating from the memorization of either the training data or the LM-pretraining corpus. 
For instance, in Table~\ref{table:hallucination_examples}, we can observe stylistic bias being injected -- ``the `dishonest' and `fake news' news outlet''.
Also, the excessive elaboration of the president's comment towards the news media, which does not appear the in source or target, can be considered informational bias -- ``they are trying to make it look like the president is trying to hide something, but it is not true!''
\textit{This analysis unveils the overlooked danger of hallucination, which is the risk of introducing political framing bias in summary generations.}
Note that this problem is not confined to MDS models only because other baseline models also have room for improvement in terms of the \feqa~hallucination score.


\noindent \paragraph{Q: What are the remaining challenges and future directions?}
The experimental results of \bartOneStep~suggest that there is room for improvement. We qualitatively checked some error cases and discovered that the title-generation is, unsurprisingly, not always accurate, and the error propagating from the title-generation step adversely affected the overall performance. Thus, one possible future direction would be to improve the neutral title generation, which would then improve the neutral summarization.

Another challenge is the subtle lexical bias involving nuanced word choices that manoeuvre readers to understand the events from biased frames. For example, ``put on hold'' and ``stalled'' both mean the same outcome, but the latter has a more negative connotations. 
Improving the model's awareness of such nuanced words or devising ways to incorporate style-transfer-based bias mitigation approaches~\cite{liu2021mitigating} could be another helpful future direction. 

We started the neutral summarization task assuming that framing bias originates from the source inputs. However, our results and analysis suggest that hallucination is another contributor to framing bias. Leveraging hallucination mitigation techniques would be a valuable future direction for the \neuS~task. 
We believe it will help to reduce informational framing bias, although it may be less effective to lexical framing biases. 
Moreover, our work can also be used to facilitate hallucination research as well. We believe the proposed framing bias metric will help researchers evaluate hallucinatory phenomena from different angles other than ``factuality''. 
The proposed framing bias metric could also be adapted to the hallucination problem without a ``neutral'' reference. The source input can substitute the ``neutral'' reference to measure if the generated summary is more politically biased than the source -- a potential indication of political hallucination.

\section{Conclusion}
\label{sec:conclusion}
We introduce a new task of Neutral Multi-News Summarization (\neuS) to mitigate media framing bias by providing a neutral summary of articles, along with the dataset \ourData~and a set of metrics. Throughout the work, we share insights to understand the challenges and future directions in the task. 
We show the relationships among polarity, extra information, and framing bias, which guides us to the metric design, while the insight that the title serves as an indicator of framing bias leads us to the model design. 
Our qualitative analysis reveals that hallucinatory content generated by models may also contribute to framing bias. 
We hope our work stimulates researchers to actively tackle political framing bias in both human-written and machine-generated texts.

\section*{Ethical Considerations}
\label{sec:ethical_considerations}



The idea of unbiased journalism has always been challenged\footnote{https://www.allsides.com/blog/does-unbiased-news-really-exist} because journalists will make their own editorial judgements that can never be guaranteed to be completely bias-free.
Therefore, we propose to generate a comprehensive summary of articles from different political leanings, instead of trying to generate a gold standard ``neutral'' article. 

One of the considerations is the bias induced by the computational approach. Automatic approaches replace a known source bias with another bias caused by human-annotated data or the machine learning models. Understanding the risk of uncontrolled adoption of such automatic tools, careful guidance should be provided in how to adopt them. For instance, an automatically generated neutral summary should be provided with reference to the original source instead of standing alone.

We use news from English-language sources only and largely American news outlets throughout this paper. Partisanship from this data refers to domestic American politics. We note that this work does not cover media bias at the international-level or in other languages. 
In future work, we will explore the application of our methodology to different cultures or languages.
However, we hope the paradigm of \neuS, providing multiple sides to neutralize the view of an issue, can encourage future research in mitigating framing bias in other languages or cultures.










\bibliography{custom}

\begin{thebibliography}{62}
\expandafter\ifx\csname natexlab\endcsname\relax\def\natexlab#1{#1}\fi

\bibitem[{all(2021)}]{allsides_2021}
 2021.
\newblock \href {https://www.allsides.com/media-bias/center} {Center -- what
  does a "center" media bias rating mean?}

\bibitem[{Bang et~al.(2021)Bang, Lee, Ishii, Madotto, and
  Fung}]{bang2021assessing}
Yejin Bang, Nayeon Lee, Etsuko Ishii, Andrea Madotto, and Pascale Fung. 2021.
\newblock Assessing political prudence of open-domain chatbots.
\newblock \emph{arXiv preprint arXiv:2106.06157}.

\bibitem[{Belinkov et~al.(2020)Belinkov, Gehrmann, and
  Pavlick}]{belinkov2020interpretability}
Yonatan Belinkov, Sebastian Gehrmann, and Ellie Pavlick. 2020.
\newblock Interpretability and analysis in neural nlp.
\newblock In \emph{Proceedings of the 58th Annual Meeting of the Association
  for Computational Linguistics: Tutorial Abstracts}, pages 1--5.

\bibitem[{Cui and Hu(2021)}]{cui2021topic}
Peng Cui and Le~Hu. 2021.
\newblock Topic-guided abstractive multi-document summarization.
\newblock In \emph{Findings of the Association for Computational Linguistics:
  EMNLP 2021}, pages 1463--1472.

\bibitem[{De~Vreese(2004)}]{de2004effects}
Claes De~Vreese. 2004.
\newblock The effects of strategic news on political cynicism, issue
  evaluations, and policy support: A two-wave experiment.
\newblock \emph{Mass Communication \& Society}, 7(2):191--214.

\bibitem[{DeYoung et~al.(2021)DeYoung, Beltagy, van Zuylen, Kuehl, and
  Wang}]{deyoung2021ms2}
Jay DeYoung, Iz~Beltagy, Madeleine van Zuylen, Bailey Kuehl, and Lucy~Lu Wang.
  2021.
\newblock Ms2: Multi-document summarization of medical studies.
\newblock \emph{arXiv preprint arXiv:2104.06486}.

\bibitem[{Durmus et~al.(2020)Durmus, He, and Diab}]{durmus-etal-2020-feqa}
Esin Durmus, He~He, and Mona Diab. 2020.
\newblock \href {https://doi.org/10.18653/v1/2020.acl-main.454} {{FEQA}: A
  question answering evaluation framework for faithfulness assessment in
  abstractive summarization}.
\newblock In \emph{Proceedings of the 58th Annual Meeting of the Association
  for Computational Linguistics}, pages 5055--5070, Online. Association for
  Computational Linguistics.

\bibitem[{Entman(1993)}]{entman1993framing}
Robert~M Entman. 1993.
\newblock Framing: Towards clarification of a fractured paradigm.
\newblock \emph{McQuail's reader in mass communication theory}, pages 390--397.

\bibitem[{Entman(2002)}]{entman2002framing}
Robert~M Entman. 2002.
\newblock Framing: Towards clarification of a fractured paradigm.
\newblock \emph{McQuail’s Reader in Mass Communication Theory. London,
  California and New Delhi: Sage}.

\bibitem[{Entman(2007)}]{entman2007framing}
Robert~M Entman. 2007.
\newblock Framing bias: Media in the distribution of power.
\newblock \emph{Journal of communication}, 57(1):163--173.

\bibitem[{Erkan and Radev(2004)}]{erkan2004lexrank}
G{\"u}nes Erkan and Dragomir~R Radev. 2004.
\newblock Lexrank: Graph-based lexical centrality as salience in text
  summarization.
\newblock \emph{Journal of artificial intelligence research}, 22:457--479.

\bibitem[{Fabbri et~al.(2019)Fabbri, Li, She, Li, and Radev}]{fabbri2019multi}
Alexander~R Fabbri, Irene Li, Tianwei She, Suyi Li, and Dragomir~R Radev. 2019.
\newblock Multi-news: A large-scale multi-document summarization dataset and
  abstractive hierarchical model.
\newblock \emph{arXiv preprint arXiv:1906.01749}.

\bibitem[{Fan et~al.(2019)Fan, White, Sharma, Su, Choubey, Huang, and
  Wang}]{fan2019plain}
Lisa Fan, Marshall White, Eva Sharma, Ruisi Su, Prafulla~Kumar Choubey, Ruihong
  Huang, and Lu~Wang. 2019.
\newblock In plain sight: Media bias through the lens of factual reporting.
\newblock \emph{arXiv preprint arXiv:1909.02670}.

\bibitem[{Gentzkow and Shapiro(2006)}]{gentzkow2006media}
Matthew Gentzkow and Jesse~M Shapiro. 2006.
\newblock Media bias and reputation.
\newblock \emph{Journal of political Economy}, 114(2):280--316.

\bibitem[{Gentzkow and Shapiro(2010)}]{gentzkow2010drives}
Matthew Gentzkow and Jesse~M Shapiro. 2010.
\newblock What drives media slant? evidence from us daily newspapers.
\newblock \emph{Econometrica}, 78(1):35--71.

\bibitem[{Gentzkow et~al.(2015)Gentzkow, Shapiro, and
  Stone}]{gentzkow2015media}
Matthew Gentzkow, Jesse~M Shapiro, and Daniel~F Stone. 2015.
\newblock Media bias in the marketplace: Theory.
\newblock In \emph{Handbook of media economics}, volume~1, pages 623--645.
  Elsevier.

\bibitem[{Goffman(1974)}]{goffman1974frame}
Erving Goffman. 1974.
\newblock \emph{Frame analysis: An essay on the organization of experience.}
\newblock Harvard University Press.

\bibitem[{Groeling(2013)}]{groeling2013media}
Tim Groeling. 2013.
\newblock Media bias by the numbers: Challenges and opportunities in the
  empirical study of partisan news.
\newblock \emph{Annual Review of Political Science}, 16:129--151.

\bibitem[{Hamborg et~al.(2019{\natexlab{a}})Hamborg, Donnay, and
  Gipp}]{hamborg2019automated}
Felix Hamborg, Karsten Donnay, and Bela Gipp. 2019{\natexlab{a}}.
\newblock Automated identification of media bias in news articles: an
  interdisciplinary literature review.
\newblock \emph{International Journal on Digital Libraries}, 20(4):391--415.

\bibitem[{Hamborg et~al.(2017)Hamborg, Meuschke, and Gipp}]{hamborg2017matrix}
Felix Hamborg, Norman Meuschke, and Bela Gipp. 2017.
\newblock Matrix-based news aggregation: exploring different news perspectives.
\newblock In \emph{2017 ACM/IEEE Joint Conference on Digital Libraries (JCDL)},
  pages 1--10. IEEE.

\bibitem[{Hamborg et~al.(2019{\natexlab{b}})Hamborg, Zhukova, and
  Gipp}]{hamborg2019illegal}
Felix Hamborg, Anastasia Zhukova, and Bela Gipp. 2019{\natexlab{b}}.
\newblock Illegal aliens or undocumented immigrants? towards the automated
  identification of bias by word choice and labeling.
\newblock In \emph{International Conference on Information}, pages 179--187.
  Springer.

\bibitem[{Hermann et~al.(2015)Hermann, Kocisky, Grefenstette, Espeholt, Kay,
  Suleyman, and Blunsom}]{hermann2015teaching}
Karl~Moritz Hermann, Tomas Kocisky, Edward Grefenstette, Lasse Espeholt, Will
  Kay, Mustafa Suleyman, and Phil Blunsom. 2015.
\newblock Teaching machines to read and comprehend.
\newblock \emph{Advances in neural information processing systems},
  28:1693--1701.

\bibitem[{Ji et~al.(2022)Ji, Lee, Frieske, Yu, Su, Xu, Ishii, Bang, Madotto,
  and Fung}]{ji2022survey}
Ziwei Ji, Nayeon Lee, Rita Frieske, Tiezheng Yu, Dan Su, Yan Xu, Etsuko Ishii,
  Yejin Bang, Andrea Madotto, and Pascale Fung. 2022.
\newblock Survey of hallucination in natural language generation.
\newblock \emph{arXiv preprint arXiv:2202.03629}.

\bibitem[{Jin et~al.(2020)Jin, Wang, and Wan}]{jin2020multi}
Hanqi Jin, Tianming Wang, and Xiaojun Wan. 2020.
\newblock Multi-granularity interaction network for extractive and abstractive
  multi-document summarization.
\newblock In \emph{Proceedings of the 58th Annual Meeting of the Association
  for Computational Linguistics}, pages 6244--6254.

\bibitem[{Laban and Hearst(2017)}]{laban2017newslens}
Philippe Laban and Marti~A Hearst. 2017.
\newblock newslens: building and visualizing long-ranging news stories.
\newblock In \emph{Proceedings of the Events and Stories in the News Workshop},
  pages 1--9.

\bibitem[{Lebanoff et~al.(2018)Lebanoff, Song, and Liu}]{lebanoff2018adapting}
Logan Lebanoff, Kaiqiang Song, and Fei Liu. 2018.
\newblock Adapting the neural encoder-decoder framework from single to
  multi-document summarization.
\newblock In \emph{Proceedings of the 2018 Conference on Empirical Methods in
  Natural Language Processing}, pages 4131--4141.

\bibitem[{Lee et~al.(2021{\natexlab{a}})Lee, Bang, Madotto, Khabsa, and
  Fung}]{lee2021towards}
Nayeon Lee, Yejin Bang, Andrea Madotto, Madian Khabsa, and Pascale Fung.
  2021{\natexlab{a}}.
\newblock Towards few-shot fact-checking via perplexity.
\newblock \emph{arXiv preprint arXiv:2103.09535}.

\bibitem[{Lee et~al.(2021{\natexlab{b}})Lee, Li, Wang, Fung, Ma, Yih, and
  Khabsa}]{lee2021unifying}
Nayeon Lee, Belinda~Z Li, Sinong Wang, Pascale Fung, Hao Ma, Wen-tau Yih, and
  Madian Khabsa. 2021{\natexlab{b}}.
\newblock On unifying misinformation detection.
\newblock \emph{arXiv preprint arXiv:2104.05243}.

\bibitem[{Lee et~al.(2020)Lee, Li, Wang, Yih, Ma, and Khabsa}]{lee2020language}
Nayeon Lee, Belinda~Z Li, Sinong Wang, Wen-tau Yih, Hao Ma, and Madian Khabsa.
  2020.
\newblock Language models as fact checkers?
\newblock \emph{arXiv preprint arXiv:2006.04102}.

\bibitem[{Lee et~al.(2019)Lee, Liu, and Fung}]{lee2019team}
Nayeon Lee, Zihan Liu, and Pascale Fung. 2019.
\newblock Team yeon-zi at semeval-2019 task 4: Hyperpartisan news detection by
  de-noising weakly-labeled data.
\newblock In \emph{Proceedings of the 13th International Workshop on Semantic
  Evaluation}, pages 1052--1056.

\bibitem[{Lee et~al.(2018)Lee, Wu, and Fung}]{lee2018improving}
Nayeon Lee, Chien-Sheng Wu, and Pascale Fung. 2018.
\newblock Improving large-scale fact-checking using decomposable attention
  models and lexical tagging.
\newblock In \emph{Proceedings of the 2018 Conference on Empirical Methods in
  Natural Language Processing}, pages 1133--1138.

\bibitem[{Lewis et~al.(2019)Lewis, Liu, Goyal, Ghazvininejad, Mohamed, Levy,
  Stoyanov, and Zettlemoyer}]{lewis2019bart}
Mike Lewis, Yinhan Liu, Naman Goyal, Marjan Ghazvininejad, Abdelrahman Mohamed,
  Omer Levy, Ves Stoyanov, and Luke Zettlemoyer. 2019.
\newblock Bart: Denoising sequence-to-sequence pre-training for natural
  language generation, translation, and comprehension.
\newblock \emph{arXiv preprint arXiv:1910.13461}.

\bibitem[{Lin(2004)}]{lin2004rouge}
Chin-Yew Lin. 2004.
\newblock Rouge: A package for automatic evaluation of summaries.
\newblock In \emph{Text summarization branches out}, pages 74--81.

\bibitem[{Liu et~al.(2018)Liu, Saleh, Pot, Goodrich, Sepassi, Kaiser, and
  Shazeer}]{liu2018generating}
Peter~J Liu, Mohammad Saleh, Etienne Pot, Ben Goodrich, Ryan Sepassi, Lukasz
  Kaiser, and Noam Shazeer. 2018.
\newblock Generating wikipedia by summarizing long sequences.
\newblock In \emph{International Conference on Learning Representations}.

\bibitem[{Liu et~al.(2021)Liu, Jia, Wei, Xu, Wang, and
  Vosoughi}]{liu2021mitigating}
Ruibo Liu, Chenyan Jia, Jason Wei, Guangxuan Xu, Lili Wang, and Soroush
  Vosoughi. 2021.
\newblock \href {http://arxiv.org/abs/2104.14795} {Mitigating political bias in
  language models through reinforced calibration}.

\bibitem[{Liu and Lapata(2019)}]{liu2019hierarchical}
Yang Liu and Mirella Lapata. 2019.
\newblock Hierarchical transformers for multi-document summarization.
\newblock In \emph{Proceedings of the 57th Annual Meeting of the Association
  for Computational Linguistics}, pages 5070--5081.

\bibitem[{Maynez et~al.(2020)Maynez, Narayan, Bohnet, and
  McDonald}]{maynez-etal-2020-faithfulness}
Joshua Maynez, Shashi Narayan, Bernd Bohnet, and Ryan McDonald. 2020.
\newblock \href {https://doi.org/10.18653/v1/2020.acl-main.173} {On
  faithfulness and factuality in abstractive summarization}.
\newblock In \emph{Proceedings of the 58th Annual Meeting of the Association
  for Computational Linguistics}, pages 1906--1919, Online. Association for
  Computational Linguistics.

\bibitem[{McCombs and Reynolds(2009)}]{mccombs2009news}
Maxwell McCombs and Amy Reynolds. 2009.
\newblock How the news shapes our civic agenda.
\newblock In \emph{Media effects}, pages 17--32. Routledge.

\bibitem[{Mohammad(2018)}]{mohammad2018obtaining}
Saif Mohammad. 2018.
\newblock Obtaining reliable human ratings of valence, arousal, and dominance
  for 20,000 english words.
\newblock In \emph{Proceedings of the 56th Annual Meeting of the Association
  for Computational Linguistics (Volume 1: Long Papers)}, pages 174--184.

\bibitem[{Morstatter et~al.(2018)Morstatter, Wu, Yavanoglu, Corman, and
  Liu}]{morstatter2018identifying}
Fred Morstatter, Liang Wu, Uraz Yavanoglu, Stephen~R Corman, and Huan Liu.
  2018.
\newblock Identifying framing bias in online news.
\newblock \emph{ACM Transactions on Social Computing}, 1(2):1--18.

\bibitem[{Nie et~al.(2019)Nie, Yao, Wang, Pan, and Lin}]{nie2019simple}
Feng Nie, Jin-Ge Yao, Jinpeng Wang, Rong Pan, and Chin-Yew Lin. 2019.
\newblock A simple recipe towards reducing hallucination in neural surface
  realisation.
\newblock In \emph{Proceedings of the 57th Annual Meeting of the Association
  for Computational Linguistics}, pages 2673--2679.

\bibitem[{Pagnoni et~al.(2021)Pagnoni, Balachandran, and
  Tsvetkov}]{pagnoni2021understanding}
Artidoro Pagnoni, Vidhisha Balachandran, and Yulia Tsvetkov. 2021.
\newblock Understanding factuality in abstractive summarization with frank: A
  benchmark for factuality metrics.
\newblock In \emph{Proceedings of the 2021 Conference of the North American
  Chapter of the Association for Computational Linguistics: Human Language
  Technologies}, pages 4812--4829.

\bibitem[{Pang et~al.(2021)Pang, Lelkes, Tran, and Yu}]{pang2021agreesum}
Richard~Yuanzhe Pang, Adam~D Lelkes, Vinh~Q Tran, and Cong Yu. 2021.
\newblock Agreesum: Agreement-oriented multi-document summarization.
\newblock \emph{arXiv preprint arXiv:2106.02278}.

\bibitem[{Papineni et~al.(2002)Papineni, Roukos, Ward, and
  Zhu}]{papineni2002bleu}
Kishore Papineni, Salim Roukos, Todd Ward, and Wei-Jing Zhu. 2002.
\newblock Bleu: a method for automatic evaluation of machine translation.
\newblock In \emph{Proceedings of the 40th annual meeting of the Association
  for Computational Linguistics}, pages 311--318.

\bibitem[{Park et~al.(2009)Park, Kang, Chung, and Song}]{park2009newscube}
Souneil Park, Seungwoo Kang, Sangyoung Chung, and Junehwa Song. 2009.
\newblock Newscube: delivering multiple aspects of news to mitigate media bias.
\newblock In \emph{Proceedings of the SIGCHI conference on human factors in
  computing systems}, pages 443--452.

\bibitem[{Park et~al.(2011)Park, Lee, and Song}]{park2011contrasting}
Souneil Park, Kyung-Soon Lee, and Junehwa Song. 2011.
\newblock Contrasting opposing views of news articles on contentious issues.
\newblock In \emph{Proceedings of the 49th annual meeting of the association
  for computational linguistics: Human language technologies}, pages 340--349.

\bibitem[{Pasunuru et~al.(2021)Pasunuru, Liu, Bansal, Ravi, and
  Dreyer}]{pasunuru2021efficiently}
Ramakanth Pasunuru, Mengwen Liu, Mohit Bansal, Sujith Ravi, and Markus Dreyer.
  2021.
\newblock Efficiently summarizing text and graph encodings of multi-document
  clusters.
\newblock In \emph{Proceedings of the 2021 Conference of the North American
  Chapter of the Association for Computational Linguistics: Human Language
  Technologies}, pages 4768--4779.

\bibitem[{Perse and Lambe(2016)}]{perse2016media}
Elizabeth~M Perse and Jennifer Lambe. 2016.
\newblock \emph{Media effects and society}.
\newblock Routledge.

\bibitem[{Raffel et~al.(2019)Raffel, Shazeer, Roberts, Lee, Narang, Matena,
  Zhou, Li, and Liu}]{raffel2019exploring}
Colin Raffel, Noam Shazeer, Adam Roberts, Katherine Lee, Sharan Narang, Michael
  Matena, Yanqi Zhou, Wei Li, and Peter~J Liu. 2019.
\newblock Exploring the limits of transfer learning with a unified text-to-text
  transformer.
\newblock \emph{arXiv preprint arXiv:1910.10683}.

\bibitem[{Recasens et~al.(2013)Recasens, Danescu-Niculescu-Mizil, and
  Jurafsky}]{recasens2013linguistic}
Marta Recasens, Cristian Danescu-Niculescu-Mizil, and Dan Jurafsky. 2013.
\newblock Linguistic models for analyzing and detecting biased language.
\newblock In \emph{Proceedings of the 51st Annual Meeting of the Association
  for Computational Linguistics (Volume 1: Long Papers)}, pages 1650--1659.

\bibitem[{Reiter(2018)}]{reiter2018structured}
Ehud Reiter. 2018.
\newblock A structured review of the validity of bleu.
\newblock \emph{Computational Linguistics}, 44(3):393--401.

\bibitem[{Sarker et~al.(2019)Sarker, Klein, Mee, Harik, and
  Gonzalez-Hernandez}]{sarker2019interpretable}
Abeed Sarker, Ari~Z Klein, Janet Mee, Polina Harik, and Graciela
  Gonzalez-Hernandez. 2019.
\newblock An interpretable natural language processing system for written
  medical examination assessment.
\newblock \emph{Journal of biomedical informatics}, 98:103268.

\bibitem[{Scheufele(2000)}]{scheufele2000agenda}
Dietram~A Scheufele. 2000.
\newblock Agenda-setting, priming, and framing revisited: Another look at
  cognitive effects of political communication.
\newblock \emph{Mass communication \& society}, 3(2-3):297--316.

\bibitem[{Scheufele and Tewksbury(2007)}]{scheufele2007framing}
Dietram~A Scheufele and David Tewksbury. 2007.
\newblock Framing, agenda setting, and priming: The evolution of three media
  effects models.
\newblock \emph{Journal of communication}, 57(1):9--20.

\bibitem[{Sides(2018)}]{sides2018media}
All Sides. 2018.
\newblock Media bias ratings.
\newblock \emph{Allsides.com}.

\bibitem[{Spinde et~al.(2021)Spinde, Kreuter, Gaissmaier, Hamborg, Gipp, and
  Giese}]{spinde2021you}
Timo Spinde, Christina Kreuter, Wolfgang Gaissmaier, Felix Hamborg, Bela Gipp,
  and Helge Giese. 2021.
\newblock Do you think it's biased? how to ask for the perception of media
  bias.
\newblock In \emph{2021 ACM/IEEE Joint Conference on Digital Libraries (JCDL)},
  pages 61--69. IEEE.

\bibitem[{Thorne et~al.(2018)Thorne, Vlachos, Christodoulopoulos, and
  Mittal}]{thorne2018fever}
James Thorne, Andreas Vlachos, Christos Christodoulopoulos, and Arpit Mittal.
  2018.
\newblock Fever: a large-scale dataset for fact extraction and verification.
\newblock \emph{arXiv preprint arXiv:1803.05355}.

\bibitem[{Wiseman et~al.(2017)Wiseman, Shieber, and
  Rush}]{wiseman-etal-2017-challenges}
Sam Wiseman, Stuart Shieber, and Alexander Rush. 2017.
\newblock \href {https://doi.org/10.18653/v1/D17-1239} {Challenges in
  data-to-document generation}.
\newblock In \emph{Proceedings of the 2017 Conference on Empirical Methods in
  Natural Language Processing}, pages 2253--2263, Copenhagen, Denmark.
  Association for Computational Linguistics.

\bibitem[{Wolf et~al.(2020)Wolf, Debut, Sanh, Chaumond, Delangue, Moi, Cistac,
  Rault, Louf, Funtowicz, Davison, Shleifer, von Platen, Ma, Jernite, Plu, Xu,
  Le~Scao, Gugger, Drame, Lhoest, and Rush}]{wolf-etal-2020-transformers}
Thomas Wolf, Lysandre Debut, Victor Sanh, Julien Chaumond, Clement Delangue,
  Anthony Moi, Pierric Cistac, Tim Rault, Remi Louf, Morgan Funtowicz, Joe
  Davison, Sam Shleifer, Patrick von Platen, Clara Ma, Yacine Jernite, Julien
  Plu, Canwen Xu, Teven Le~Scao, Sylvain Gugger, Mariama Drame, Quentin Lhoest,
  and Alexander Rush. 2020.
\newblock \href {https://doi.org/10.18653/v1/2020.emnlp-demos.6} {Transformers:
  State-of-the-art natural language processing}.
\newblock In \emph{Proceedings of the 2020 Conference on Empirical Methods in
  Natural Language Processing: System Demonstrations}, pages 38--45, Online.
  Association for Computational Linguistics.

\bibitem[{Yano et~al.(2010)Yano, Resnik, and Smith}]{yano2010shedding}
Tae Yano, Philip Resnik, and Noah~A Smith. 2010.
\newblock Shedding (a thousand points of) light on biased language.
\newblock In \emph{Proceedings of the NAACL HLT 2010 Workshop on Creating
  Speech and Language Data with Amazon’s Mechanical Turk}, pages 152--158.

\bibitem[{Zhang et~al.(2019{\natexlab{a}})Zhang, Zhao, Saleh, and
  Liu}]{zhang2019pegasus}
Jingqing Zhang, Yao Zhao, Mohammad Saleh, and Peter~J. Liu. 2019{\natexlab{a}}.
\newblock \href {http://arxiv.org/abs/1912.08777} {Pegasus: Pre-training with
  extracted gap-sentences for abstractive summarization}.

\bibitem[{Zhang et~al.(2019{\natexlab{b}})Zhang, Da~San~Martino,
  Barr{\'o}n-Cedeno, Romeo, An, Kwak, Staykovski, Jaradat, Karadzhov, Baly
  et~al.}]{zhang2019tanbih}
Yifan Zhang, Giovanni Da~San~Martino, Alberto Barr{\'o}n-Cedeno, Salvatore
  Romeo, Jisun An, Haewoon Kwak, Todor Staykovski, Israa Jaradat, Georgi
  Karadzhov, Ramy Baly, et~al. 2019{\natexlab{b}}.
\newblock Tanbih: Get to know what you are reading.
\newblock \emph{EMNLP-IJCNLP 2019}, page 223.

\end{thebibliography}
\bibliographystyle{acl_natbib}

\appendix
\section*{Appendix} 

\section{Topics covered in \ourData dataset} 
The \ourData dataset language is English and mainly focuses on U.S. political topics that often result in media bias. The top-5 most frequent topics are `Elections', `White House', `Politics', `Coronavirus', `Immigration'. 

The full list is as follow (in a descending order of frequency): 
[`Elections', `White House', `Politics', `Coronavirus', `Immigration', `Violence in America', `Economy and Jobs', `Supreme Court', `Middle East', `US House', `Healthcare', `World', `US Senate', `National Security', `Gun Control and Gun Rights', `Media Bias', `Federal Budget', `Terrorism', `US Congress', `Foreign Policy', `Criminal Justice', `Justice Department', `Trade', `Impeachment', `Donald Trump', `North Korea', `Russia', `Education', `Environment', `Free Speech', `FBI', nan, `Abortion', `General News', `Disaster', `US Military', `Technology', `LGBT Rights', `Sexual Misconduct', `Voting Rights and Voter Fraud', `Joe Biden', `Race and Racism', `Economic Policy', `Justice', `Holidays', `Taxes', `China', `Polarization', `Democratic Party', `Religion and Faith', `Sports', `Homeland Security', `Culture', `Cybersecurity', `National Defense', `Public Health', `Civil Rights', `Europe', `Great Britain', `Banking and Finance', `Republican Party', `NSA', `Business', `State Department', `Facts and Fact Checking', `Media Industry', `Labor', `Veterans Affairs', `Campaign Finance', `Life During COVID-19', `Transportation', `Marijuana Legalization', `Agriculture', `Arts and Entertainment', `Fake News', `Campaign Rhetoric', `Nuclear Weapons', `Israel', `Asia', `CIA', `Role of Government', `George Floyd Protests', "Women's Issues", `Safety and Sanity During COVID-19', `Animal Welfare', `Treasury', `Science', `Climate Change', `Domestic Policy', `Energy', `Housing and Homelessness', `Bridging Divides', `Mexico', `Inequality', `COVID-19 Misinformation', `ISIS', `Palestine', `Bernie Sanders', `Tulsi Gabbard', `Sustainability', `Family and Marriage', `Pete Buttigieg', `Welfare', `Opioid Crisis', `Amy Klobuchar', `Food', `EPA', `South Korea', `Alaska: US Senate 2014', `Social Security', `US Constitution', `Tom Steyer', `Andrew Yang', `Africa']

\section{Additional Salient Information Score Results}
We report additional Salient information F1 (Table \ref{table:rouge_scores_f1}) and Recall (Table \ref{table:rouge_scores_recall}) scores for {\scshape Rouge1}, {\scshape Rouge2} and {\scshape RougeL}. 

\begin{table}[h]
\centering
\resizebox{\linewidth}{!}{
\begin{tabular}{lccc}
\toprule
 & \begin{tabular}[c]{@{}c@{}}{\scshape Rouge1} \\ {\scshape F1}\end{tabular} & \begin{tabular}[c]{@{}c@{}}{\scshape Rouge2} \\ {\scshape F1}\end{tabular} & \begin{tabular}[c]{@{}c@{}}{\scshape RougeL} \\ {\scshape F1}\end{tabular} \\ \midrule
\lexrank & 33.60\% & 13.60\% & 29.77\% \\
\bartCNN & 33.76\% & 13.67\% & 30.57\% \\
\pegasusMulti & 30.03\% & 10.28\% & 26.70\% \\
\bartMulti & 23.01\% & 6.84\% & 20.55\% \\
\bartNaive & 36.76\% & 16.27\% & 32.86\% \\ \midrule
\bartOneStep & 35.49\% & 15.69\% & 32.05\% \\ \bottomrule
\end{tabular}}
\caption{Additional Salient Info Scores. F1 scores for {\scshape Rouge1}, {\scshape Rouge2} and {\scshape RougeL} for \ourData~testset. For the scores, the \textit{higher} number is the better.}
\label{table:rouge_scores_f1}
\end{table}

\begin{table}[h]
\centering
\resizebox{\linewidth}{!}{
\begin{tabular}{lccc}
\toprule
 & \begin{tabular}[c]{@{}c@{}}{\scshape Rouge1} \\ {\scshape Recall}\end{tabular} & \begin{tabular}[c]{@{}c@{}}{\scshape Rouge2} \\ {\scshape Recall}\end{tabular} & \begin{tabular}[c]{@{}c@{}}{\scshape RougeL} \\ {\scshape Recall}\end{tabular} \\ \midrule
\lexrank & 39.08\% & 17.66\% & 34.69\% \\
\bartCNN & 35.63\% & 15.32\% & 32.22\% \\
\pegasusMulti & 44.42\% & 16.99\% & 39.45\% \\
\bartMulti & 35.76\% & 12.48\% & 32.08\% \\
\bartNaive & 35.11\% & 15.74\% & 31.43\% \\ \midrule
\bartOneStep & 36.07\% & 16.47\% & 32.63\% \\ \bottomrule
\end{tabular}}
\caption{Additional Salient Info Scores. Recall scores for {\scshape Rouge1}, {\scshape Rouge2} and {\scshape RougeL} for \ourData~testset. For the scores, the \textit{higher} number is the better.}
\label{table:rouge_scores_recall}
\end{table}


\section{Details for Human Evaluation (A/B testing)}

We first presented the participants with the definition of framing bias from our paper, and also showed examples in Table 1 to ensure they understand what framing bias is. Then we asked the following question: ``Which one of the articles do you believe to be more biased toward one side or the other side in the reporting of news?'' This is modified to serve as a question for AB testing based on ``To what extent do you believe that the article is biased toward one side or the other side in the reporting of news?'' The original question is one of the 21 questions which are suitable and reliable for measuring the perception of media bias, designed by \citet{spinde2021you}.

The participants (research graudate students) have different nationalities including Canada, China, Indonesia, Iran, Italy, Japan, Poland and South Korea (ordered in an alphabetical order).  All of participants answered to be not having political leaning towards U.S. politics. All participants are fully explained on the usage of collected data in this particular work and agreed on it.


\section{Experimental Setup Details}
All our experimental codes are based on the HuggingFace~\cite{wolf-etal-2020-transformers}. We used the following hyperparameters during training and across models: $10$ epoch size, $3e-5$ learning rate, and a batch size of $16$. We did not do hyper-parameters tuning since our objective is to provide various baselines and analysis. Training run-time for all of our experiments are fast ($<6$hr). We ran all experiments with one NVIDIA 2080Ti GPU with 16 GB of memory. The experiment was a single-run.

\section{Generation Examples from Different Models}
To help better understand performances of each models, we provide more examples of generation from all baseline models and our proposed models along with the target neutral article summary. The examples can be found in Table \ref{appendix_table:generation_examples}, \ref{appendix_table:generation_examples2}, \ref{appendix_table:generation_examples3}. 

\section{Illustration of hallucinatory framing bias}
In Table \ref{appendix_table:hallu_generation_examples}, we provide more examples of hallucinatory framing bias that was discussed in analysis.

\label{sec:appendix_understanding_framing_bias}

\begin{table*}
\centering
\small
\resizebox{\linewidth}{!}{
    \centering
    \begin{tabular}{p{15cm}}
    \toprule
    \textit{\textbf{Target:} The House Ways and Means Committee has requested that the IRS turn over the last six years of tax returns from President Trump.}
    \\ \midrule\midrule
    \textbf{\lexrank:} Representative Richard Neal invoked law that allows House Ways and Means Committee chair to obtain any person’s tax information. The request by House Democrats for six years of Donald Trump's tax returns opens a serious new \textcolor{black}{political}, legal, constitutional and personal front in the party's \textcolor{black}{battle} to use its new majority to check a \textcolor{black}{defiant} President.
    \\ \midrule
    \textbf{\bartCNN:} Rep. Richard Neal invoked law that allows House Ways and Means Committee chair to obtain any person’s tax information. The request by House Democrats for six years of Donald Trump's tax returns opens a serious new political, legal, constitutional and personal front in the party's battle to use its new majority to check a defiant President.
    \\ \midrule
    \textbf{\pegasusMulti:} The battle over President Trump's tax returns is officially on. Rep. Richard Neal, chairman of the House Ways and Means Committee, on Wednesday formally asked the IRS for six years of Trump's personal and business tax returns, as well as the returns of some of his companies, reports the Washington Post. The panel has the power to subpoena someone's tax information, and Neal has signaled that he plans to do just that. "The American people have a right to know who the president of the United States is, and we have a responsibility to do our part to ensure that the tax laws are administered in an efficient and thorough manner," he said in a statement, per the Hill. The White House has signaled that it will fight the request, with press secretary Sarah Huckabee Sanders calling it "an abuse of the tax-writing committee's authority." The letter to the IRS was signed by Neal, a Massachusetts Democrat, and was obtained by the Wall Street Journal. The Journal notes that Neal has the authority under a law passed in 1969 that allows the House Ways and Means Committee to obtain tax information from anyone. 
    \\ \midrule
    \textbf{\bartMulti:} the head of the powerful house tax-writing committee has asked the government for six years of president trump's tax returns, a move that is expected to lead to a long court battle with the white house, the washington post reports. in a letter to the irs on wednesday, rep. richard h. Neal, chairman of the house ways and means committee, invoked a law that allows any person to obtain any person's personal tax information, the wall street journal reports. according to the hill, the move is a " serious new political, legal, constitutional, and personal front in the party's battle to use its new majority to check a defiant president. " it is a bid to solve one of the most tantalizing and enduring mysteries of the trump campaign and his presidency — what is it in the president's unseen financial and business life that he doesn't want americans to see?
    \\ \midrule
    \textbf{\bartNaive:} House Ways and Means Committee Chair Richard Neal (D-MA) has formally requested President Trump's tax returns from the IRS.
    \\ \midrule
    \textbf{\bartOneStep:} TITLE=> House Committee Seeks Trump Tax Returns. ARTICLE=> The House Ways and Means Committee formally requested President Trump’s tax returns on Wednesday, kicking off what could be a protracted legal fight between Congress and the Trump administration.
    \\ \bottomrule
\end{tabular}
}
\caption{Neutral Summary Generation Examples from baseline models and \bartOneStep.}
\label{appendix_table:generation_examples}
\end{table*}

\begin{table*}
\centering
\small
\resizebox{\linewidth}{!}{
    \centering
    \begin{tabular}{p{15cm}}
    \toprule
    \textit{\textbf{Target:} The State Department has blocked Gordon Sondland, the U.S. Ambassador to the European Union, from testifying at a scheduled deposition on the Ukraine controversy. Sondland is a key figure as he had texted with Bill Taylor, a top U.S. diplomat in Ukraine, arguing over allegations of a quid pro quo. This move by the State Department has triggered a new clash with House Democrats as part of their impeachment inquiry, and President Trump has dismissed the deposition as a ``a kangaroo court'' as he defended the State Department decision.}
    \\ \midrule \midrule
    \textbf{\lexrank:} The Trump administration on Tuesday blocked a planned deposition from Gordon Sondland, the U.S. ambassador to the European Union and a central figure in the Democrats’ impeachment inquiry, according to a statement by his lawyer. Democrats swiftly called the administration’s move an act of \textcolor{black}{obstruction} and \textcolor{black}{accused} President Trump of \textcolor{black}{hiding} information about his efforts to pressure the Ukrainian government to investigate his \textcolor{black}{political} rivals.
    \\ \midrule 
    \textbf{\bartCNN:} The State Department instructed Ambassador to the European Union Gordon Sondland not to appear for the deposition, according to his counsel. Democrats swiftly called the administration’s move an act of \textcolor{black}{obstruction} and \textcolor{black}{accused} President Trump of \textcolor{black}{hiding} information about his efforts to pressure the Ukrainian government to investigate his \textcolor{black}{political} rivals.
    
    \\ \midrule 
    \textbf{\pegasusMulti:} The Trump administration has blocked US ambassador to the European Union Gordon Sondland from testifying before a House panel investigating possible obstruction of justice by the president, the New York Times reports. Sondland is a key witness in the probe into whether Trump obstructed justice when he fired FBI chief James Comey. Sondland had been scheduled to testify in front of the House Judiciary Committee on Tuesday, but the State Department ordered him not to appear, according to the Washington Post. Sondland is a former British ambassador to the US. In a Tuesday tweet, President Trump said he would love to send Sondland to testify, "but unfortunately he would be testifying before a totally compromised kangaroo court, where Republican's rights have been taken away, and true facts are not allowed out for the public to see." He added: "I believe you are incorrect about President Trump's intentions."

    \\ \midrule
    \textbf{\bartMulti:} a key witness in the house impeachment inquiry was ordered not to appear at a scheduled deposition on tuesday, ramping up tensions between congressional democrats and the trump administration. the state department instructed ambassador to the eu and key figure in the impeachment inquiry, Gordon sondland, not to show up for the deposition, according to his counsel. democrats called the move an act of obstruction and accused president trump of hiding information about his efforts to pressure the ukraine government to investigate his political rivals, the new york times reports. " this is a clear attempt to block testimony from key witnesses in the ongoing impeachment investigation, " said house minority leader nancy pelosi, per the washington post. " it's an impeachable act. " in a tweet tuesday night, trump said he " would love to send ambassador [ sundland ], a really good man and great american, to testify, but unfortunately he would be testifying before a totally compromised kangaroo court, where republican ’ s rights have been taken away, and true facts are not allowed out for the public to see. "
    \\ \midrule
    \textbf{\bartNaive:} The Trump administration blocked a planned deposition from Gordon Sondland, the U.S. ambassador to the European Union and a central figure in the Democrats’ impeachment inquiry, according to a statement from his lawyer. President Trump responded via Twitter, saying that he ``would love to send Ambassador Sonderland, a really good man and great American, to testify, but \textcolor{black}{unfortunately} he would be testifying before a totally compromised kangaroo court, where Republican’s rights have been taken away, and true facts are not allowed out for the public to see.''
    \\ \midrule 
    \textbf{\bartOneStep:} TITLE=> State Dept Orders Ambassador Sondland Not to Testify Before House. ARTICLE=> The State Department has ordered Ambassador Gordon Stondland not to testify before the House of Representatives in the impeachment inquiry of President Donald Trump. Democrats have \textcolor{black}{accused} the Trump administration of blocking statements from witnesses.
    \\ \bottomrule
    
\end{tabular}
}
\caption{Continued from Previous Page: Neutral Summary Generation Examples from baseline models and \bartOneStep.}
\label{appendix_table:generation_examples2}
\end{table*}

\begin{table*}
\centering
\small
\resizebox{\linewidth}{!}{
    \centering
    \begin{tabular}{p{15cm}}
    \toprule
    \textit{\textbf{Target:} Ukrainian police have opened an investigation into whether or not U.S. Ambassador Marie Yovanovitch came under surveillance before she was recalled from her post in Ukraine last May. Democrats have released documents that show Lev Parnas, an associate of Rudy Giuliani, communicating about Yovanovitch's removal.}
    \\ \midrule\midrule
    \textbf{\lexrank:} Ukraine’s government announced Thursday that police are investigating whether ousted U.S. ambassador Marie Yovanovitch was subject to \textcolor{black}{illegal} surveillance, in response to new documents released ahead of President Trump’s \textcolor{black}{impeachment} trial. Those documents, released by Democratic lawmakers, showed Lev Parnas -- an associate of Trump lawyer Rudy Giuliani -- communicating about the removal of Marie Yovanovitch as the ambassador to Ukraine.
    \\ \midrule
    \textbf{\bartCNN:} Police in Ukraine have opened a criminal investigation into whether allies of President Trump had the U.S. ambassador to the country under surveillance. Newly released documents suggest that Ms. Yovanovitch was being watched in Kyiv. The Ministry of Internal Affairs of Ukraine says it's asking the FBI to help its inquiry.
    \\ \midrule
    \textbf{\pegasusMulti:} The Ukraine has opened a criminal investigation into allegations that President Trump's allies had the US ambassador to the country under surveillance while she was stationed in the country, the New York Times reports. According to the Wall Street Journal, the allegations surfaced in newly released documents released by Democrats ahead of Trump's impeachment trial. The documents show that Rudy Giuliani's associate, Lev Parnas, communicated about the removal of Marie Yovanovitch as ambassador to Ukraine with Yovanovitch's former boss at the State Department, Sergey Kislyak. Yovanovitch was removed from her post in May of last year. Ukraine's Ministry of Internal Affairs says it's asking the FBI to help with the investigation. "Ukraine cannot ignore such illegal activities on the territory of its own state," the ministry says in a statement
    \\ \midrule
    \textbf{\bartMulti:} new documents released ahead of president trump's impeachment trial suggest that the us ambassador to the country was under surveillance while she was stationed in the country, the new york times reports. according to the wall street journal, the allegations were made in a text message between a us politician and an associate of trump lawyer rudy giuliani. in the messages, the politician says he'd like to see former ambassador to ukraine and current ambassador to russia, former deputy foreign minister oleksandr turchynov, removed from his post. " i would like to remove him from his position, " the politician wrote in the message, which was sent in april 2017. the messages were sent in response to newly released documents from the house intelligence committee. those documents show that former trump lawyer paul ryan had been trying to get rid of yovanovitch, who was ambassador at the time, for more than a year, the journal reports. the messages also show that ryan's chief of staff, george w. bush, had been in touch with ryan about removing her from her post. 
    \\ \midrule
    \textbf{\bartNaive:} Ukraine's national police are investigating allegations that U.S. Ambassador Marie Yovanovitch was \textcolor{black}{subjected} to \textcolor{black}{illegal} surveillance while she was stationed in Kyiv.
    \\ \midrule
    \textbf{\bartOneStep:} TITLE=> Ukraine Opens Investigation Into Claims U.S. Ambassador Yovanovitch Was Under Surveillance. ARTICLE=> Ukraine's national police are investigating allegations that former U. S. ambassador Marie Yovanovich was \textcolor{black}{subjected} to surveillance while she was stationed in Kyiv, following the release of documents by Democratic lawmakers.
    \\ \bottomrule
\end{tabular}
}
\caption{Continued from Previous Page: Neutral Summary Generation Examples from baseline models and \bartOneStep.}
\label{appendix_table:generation_examples3}
\end{table*}

\begin{table*}[t]
\centering
\small
\resizebox{\linewidth}{!}{
    \centering
    \begin{tabular}{p{15cm}}
    \toprule
    \textbf{[1] Source:} Newly released documents suggest that Ms. Yovanovitch was being watched in Kyiv while she was the U.S. ambassador there. The police in Ukraine have opened a criminal investigation into whether allies of President Trump had the United States ambassador to the country under surveillance while she was stationed in Kyiv, the Ukrainian government said on Thursday. [SEP] Ukraine’s government announced Thursday that police are investigating whether ousted U.S. ambassador Marie Yovanovitch was subject to illegal surveillance, in response to new documents released ahead of President Trump’s impeachment trial. Those documents, released by Democratic lawmakers, showed Lev Parnas -- an associate of Trump lawyer Rudy Giuliani -- communicating about the removal of Marie Yovanovitch as the ambassador to Ukraine. [SEP] Ukraine's national police are investigating allegations that U.S. Ambassador Marie Yovanovitch was under surveillance in Kyiv last spring — an allegation made in WhatsApp messages between a U.S. politician and Lev Parnas. ``Ukraine cannot ignore such illegal activities on the territory of its own state,'' the Ministry of Internal Affairs of Ukraine said in a statement Thursday. The ministry says it's asking the FBI to help its inquiry.
    \\ \midrule
    \textbf{[1] MDS Hallucination:} new documents released ahead of president trump's impeachment trial suggest that the us ambassador to the country was under surveillance while she was stationed in the country, the new york times reports. according to the wall street journal, the allegations were made in a text message between a us politician and an associate of trump lawyer rudy giuliani. in the messages, the politician says he'd like to see former ambassador to ukraine and current ambassador to russia, former deputy foreign minister oleksandr turchynov, removed from his post. ``i would like to remove him from his position,'' the politician wrote in the message, which was sent in april 2017. the messages were sent in response to newly released documents from the house intelligence committee. those documents show that former trump lawyer paul ryan had been trying to get rid of yovanovitch, who was ambassador at the time, for more than a year, the journal reports. the messages also show that ryan's chief of staff, george w. bush, had been in touch with ryan about removing her from her post. 
    \\ \bottomrule \\\toprule
    \textbf{[2] Source:}  ``If I threaten someone, you’ll know it,'' White House counselor Kellyanne Conway told reporters Friday morning in a bizarre hours-long effort to discredit an audio recording that seemingly showed her threatening to use West Wing resources to investigate the personal life of a reporter a day earlier. ``I never threatened anyone,'' Conway told a gaggle of journalists at the White House. ``Don’t use those words. I didn’t threaten. It’s not a threat. I never threatened her.'' [SEP] A tense phone conversation between a reporter for the Washington Examiner and White House senior counselor Kellyanne Conway was published by the newspaper on Thursday. In the conversation, Conway objected that a story written by the reporter, Caitlin Yilek, mentioned that her husband George Conway is a fierce critic of President Trump on Twitter. Yilek was writing a story on Conway possibly becoming President Trump's next White House chief of staff if Trump decides to move on from the official now in the position, Mick Mulvaney. [SEP] Kellyanne Conway defended her phone call with a Washington Examiner reporter during an appearance on Fox News. Fox News host Sandra Smith described the phone call Friday between reporter Caitlin Yilek and the counselor to the president as a "heated clash" during which things got ``very personal.'' Conway claimed, ``Everything I said in that phone call, I’ve said publicly before. Every single thing. I said it again.'' Smith asked Conway if there was anything she regretted about how she handled her phone call, which she denied.
    \\ \midrule
    \textbf{[2] MDS Hallucination:} a recording of a tense phone call between senior white house adviser and reporter catherine yilek was published by the washington examiner on friday, and it's not going over well. in the recording, yalek says she was told by the white house that she would be fired if she didn't back off from her reporting on president trump's controversial son-in-law, and that she'd have to go to court if she did not. ``i'm not going to do that,'' she says in the recorded conversation, per the new york times. ``you're going to get fired if you don't shut the f --- up.'' in the call, she also says she'll use the office of management and budget to investigate the personal life of the reporter. ``if i threaten someone, you'll know it,'' the caller can be heard saying in the audio recording, per politico. ``don ’ t use those words. it ’ s not a threat. i never threatened anyone.'' but on monday, white house counselor to the president katie holmes told fox news that she had never threatened the reporter.
\\ \bottomrule
\end{tabular}
}
\caption{Examples of hallucinatory framing bias from MDS models and the corresponding the source input.}
\label{appendix_table:hallu_generation_examples}
\end{table*}


\end{document}